\documentclass[letterpaper, 10 pt, conference]{ieeeconf}  

\IEEEoverridecommandlockouts                              

\usepackage{amsmath}
\usepackage{amssymb}
\usepackage{booktabs}
\usepackage{multirow}
\usepackage{gensymb}
\let\labelindent\relax
\usepackage{enumitem}
\usepackage{pifont}
\usepackage{graphicx}
\usepackage{xcolor}
\usepackage[pagebackref,breaklinks,colorlinks]{hyperref}

\usepackage{siunitx}
\usepackage{balance} 
\usepackage{ulem}
\usepackage{soul}
\usepackage{float}
\usepackage{hyperref}
\usepackage{multirow}
\usepackage{xcolor,colortbl}
\definecolor{ashgrey}{rgb}{0.7, 0.75, 0.71}
\definecolor{gainsboro}{rgb}{0.86, 0.86, 0.86}

\newcommand{\erw}[1]{\textcolor[rgb]{0.00, 0.00, 0.00}{#1}}
\newcommand{\jac}[1]{\textcolor[rgb]{0.00, 0.00, 0.00}{#1}}
\newcommand{\lae}[1]{\textcolor[rgb]{0.00, 0.00, 0.00}{#1}}
\newcommand{\lastmodif}[1]{\textcolor[rgb]{0.00, 0.00, 0.00}{#1}}
\newcommand{\matr}[1]{\mathbf{#1}}

\title{\LARGE \bf Attention Graph for Multi-Robot Social Navigation with Deep Reinforcement Learning}

\author{Erwan Escudie$^{1}$, Laetitia Matignon$^{2}$ and Jacques Saraydaryan$^{3}$ \\ Project Page: \href{https://perso.liris.cnrs.fr/laetitia.matignon/multisoc.html}{https://perso.liris.cnrs.fr/laetitia.matignon/multisoc.html}
\thanks{$^{1}$Erwan Escudie is with Univ Lyon, LIRIS, UMR5205, CITI Lab., INRIA-INSA Chroma team, Villeurbanne, France} %
\thanks{$^{2}$Laetitia Matignon is with Univ Lyon, UCBL, CNRS, INSA Lyon, LIRIS, UMR5205, F-69622, France. {\tt\small laetitia.matignon@univ-lyon1.fr}}%
\thanks{$^{3}$Jacques Saraydaryan is with CPE Lyon, CITI Lab.,  INRIA-INSA Chroma team, Villeurbanne, France. {\tt\small jacques.saraydaryan@cpe.fr}
}
}

\begin{document}

\maketitle
\thispagestyle{empty}
\pagestyle{empty}

\begin{abstract}
Learning robot navigation strategies among pedestrian is crucial for domain based applications. Combining perception, planning and prediction allows us to model the interactions between robots and pedestrians, resulting in impressive outcomes especially with recent approaches based on deep reinforcement learning (RL). However, these works do not consider multi-robot scenarios. In this paper, we present MultiSoc, a new method for learning multi-agent socially aware navigation strategies using RL. Inspired by recent works on multi-agent deep RL, our method leverages graph-based representation of agent interactions, combining the positions and fields of view of entities (pedestrians and agents). Each agent uses a model based on two Graph Neural Network combined with attention mechanisms. First an edge-selector produces a sparse graph, then a crowd coordinator applies node attention to produce a graph representing the influence of each entity on the others. This is incorporated into a model-free RL framework to learn multi-agent policies. We evaluate our approach on simulation and provide a series of experiments in a set of various conditions (number of agents / pedestrians). Empirical results show that our method learns faster than social navigation deep RL mono-agent techniques, and enables efficient multi-agent implicit coordination in challenging crowd navigation with multiple heterogeneous humans. Furthermore, by incorporating customizable meta-parameters, we can adjust the neighborhood density to take into account in our navigation strategy.
\end{abstract}

\section{Introduction} 

 \begin{figure}[t]
    \centering
    \includegraphics[width=1.0\linewidth]{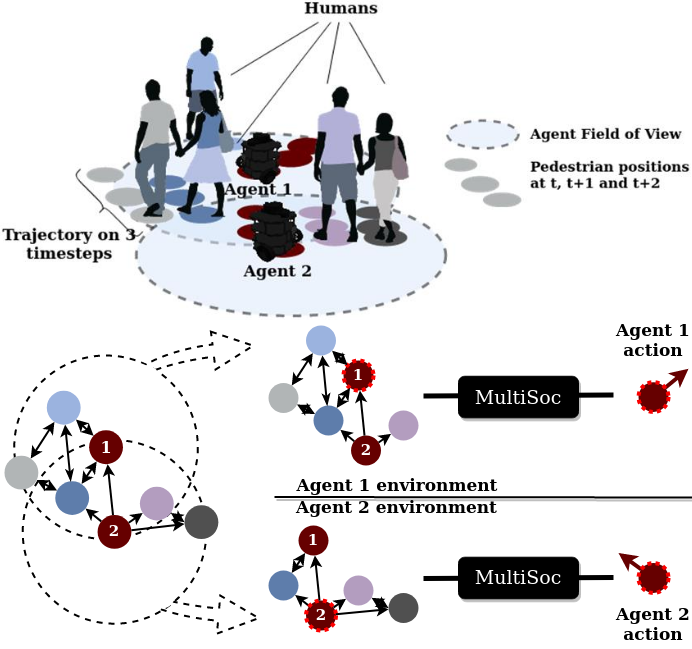}
    \caption{Overview of MultiSoc process: (Top) Scene with two agents (robots) with 360° field of view (FoV). (Bottom) Each agent applies MultiSoc on a graph of its environment (limited to its FoV) with each entities (human/robot) as a node.}
    \label{fig:overview}
\end{figure}

  \lae{
Robot navigation in crowded spaces has attracted significant attention in recent years given its numerous potential applications, but it still faces many challenges \cite{mavrogiannis_core_2023}. Especially understanding pedestrian behavior is crucial to develop effective robot navigation strategies that prioritize human safety. But predicting crowd behavior is difficult and most of approaches intend to learn it from experiment or simulation \cite{Han2020, huang2022gst, Zhou2014LearningCC}. Robot social navigation becomes even more challenging when the crowd is dense or the environment complex (obstacles, occlusions, reduced field of view, ...). 
Another difficulty appears when multiple robots navigate in the same crowd. Then a distinction must be made between a robot or human from the navigation point of view, as robots can coordinate themselves.
}

\lae{
Recent approaches \cite{chen2020gcn_gaze,liu2021dsrnn} use deep reinforcement learning (RL) to build social navigation strategies with the help of a simulated crowd. Lately the works of \cite{liu2023attngraph} use deep RL combined with attention \cite{vaswani2017attention} and graph-based representations \cite{Scarselli:2009ku} of interactions between robot and humans. This achieved very good performance in dense crowds for a single robot.  However, as specified by the authors, this model remains difficult to train as it exhibits unstable learning. Furthermore, its architecture explicitly separates the robot from humans and, as such, cannot easily be extended in its current form to accommodate multi-robot learning. }

\lae{
In this work, we propose a model for the learning of multi-robot navigation strategies within crowded environments (cf. Fig.\ref{fig:overview} (Top)). Main challenges compared to the state of the art are learning coordinated and human-safe navigation strategies for the fleet of robots and managing interactions with both controlled entities (robots) and uncontrolled entities (humans). In our contribution, we highlight the similarity between two approaches that utilize Graph Neural Networks (GNN) \cite{Scarselli:2009ku} to represent, on one hand, human interactions in single-robot social navigation \cite{liu2023attngraph}, and on the other hand, interactions between agents in multi-robot navigation \cite{yang2023magex}. Thus GNNs offer a bridge between these two fields which we leverage in our proposed model, named MultiSoc. MultiSoc uses two GNNs combined with attention mechanisms. First an edge-selector produces a sparse graph of the most interesting interactions between entities; then a crowd coordinator applies node attention to produce a graph representing
the influence of each entity on the others. MultiSoc follows Centralized Training Decentralized Execution (CTDE) paradigm. During the learning process with a multi-agent RL algorithm, the model is shared between robots, taking benefit of each robot experience. But at the execution, each robot processes its input through its MultiSoc model and gets as result its action (commands in velocity) (cf. Fig. \ref{fig:overview} (Bottom)). The input is a directed graph with information (current and predicted future poses) concerning the entities (robots and humans) in the field of view (FoV) of the robot.
}

\lae{
We present empirical results obtained with a multi-agent social navigation simulator that we implemented building upon an existing single-agent one \cite{liu2023attngraph}. 
Our MultiSoc model overcomes the main baseline in deep RL social navigation especially when several robots are involved.  
Results also demonstrate (i) a better generalization of our model even in more balanced crowd (as many robots as humans) or with heterogeneous human policies; (ii) scalability capacities \lastmodif{with different proportions of humans and robots  between training and testing;}
(iii) the usefulness of the density factor we introduced to adapt to the neighborhood density.
}

\lae{
The main contributions of this paper are as follows. (i) We propose the first (as far as we know) graph-based interactions model for \textbf{multi}-robot social navigation (ii) We introduce a customizable meta-parameter to \textbf{adjust the neighborhood density} to take into account in each robot navigation strategy (iii) The experiments demonstrate that our model enables \textbf{efficient multi-agent implicit coordination} in challenging crowd navigation and is able to deal with \textbf{heterogeneous human policies}. 
}

\section{Related Works}
\subsection{Social robot navigation}

Social robot navigation has inspired a significant amount of research \cite{mavrogiannis_core_2023}. Early methods \cite{SFM,ORCA,berg2008} consider humans as dynamic but non-responsive obstacles resulting in shortsighted and unnatural robot behaviors.
Others plan robot motion conditioned on the predicted future trajectories of humans 
but suffer from the "freezing robot problem" \cite{Trautman2010}.

One solution is to couple planning and prediction but these suffer from computational intractability \cite{mavrogiannis_core_2023}.

A set of recent approaches address these coupled models with \textbf{deep RL}. Efficient robot policies can be trained with interaction awareness encoded in the reward function. The complexity of the coupled models is then transferred in the training. 
These approaches differ in their way of modeling crowd interactions. Previous ones ignore human-human (H-H) interactions and consider a limited number of robot-human (R-H) interactions \cite{Chen2017SACADRL,GA3C-CADRL}. Thus their performance degrades in dense and highly interactive crowds. Most of the following works use \textbf{graph-based models} to extract efficient representations of the crowd, and attention mechanisms \cite{vemula2018social} to infer the relative importance of each interaction. 
In the graph representation, the nodes represent robot or humans and the edges relations between them. 
DS-RNN \cite{liu2021dsrnn} uses a spatio-temporal graph to capture spatial R-H interactions and temporal interactions in the robot's own trajectory. Spatial and temporal interactions are represented with Recurrent Neural Networks (RNN) and attention weights are assigned to spatial edges to infer the most pertinent human neighbors. However, the number of humans is fixed for each learnt model and H-H interactions are not considered. Besides, it suffers dramatic loss when the robot encounters too many kind of obstacles.

Others employed GNNs to learn interactions between the entities. RGL \cite{chen2020relational} combines Graph Convolutional Networks (GCN) with relational graph learning but does not distinguish between H-H and R-H interactions. G-GCNRL \cite{chen2020gcn_gaze} learns human-like attention weights with a GCN trained with human gaze data.

These weights are incorporated into the adjacency matrix of a second GCN containing the policy network of the RL architecture. Lately, Attention-Based Interaction Graph (AttnGraph) \cite{liu2023attngraph} emphasizes crowd analysis by prioritizing H-H interactions. Attention mechanisms based on Graph Attention Network (GAT) are applied, first between humans to balance each human trajectory with the others; secondly to include the robot in the analysis. It is interesting to note that this method uses human trajectory predictors, some of which are also based on GNNs \cite{huang2019stgat,huang2022gst}.

\lae{Although the impressive results of AttnGraph compared to other deep RL models in social robot navigation, it is worth mentioning, according to the authors, that it remains difficult to train. Moreover, no approaches consider \textbf{multi-robot} navigation \lastmodif{with deep multi-agent RL} to our knowledge. On its part, AttnGraph architecture explicitly separates the robot from humans and cannot easily be extended to accommodate multi-robot learning.}

\subsection{Multi-agent deep reinforcement learning}

Significant developments have been made in the field of multi-agent deep RL, analysed in recent surveys \cite{Wei2021deeprl,Gronauer22deeprl}. In our context of multi-robot navigation in a crowd, we will focus solely on cooperative approaches. 
Some solutions are based on value decomposition in CTDE paradigm, as in \cite{sunehag2018vdn} where global Q-value is decomposed based on individual Q-values. QMIX \cite{rashid2020qmix} adds a constraint of monotonicity on the type of function used to merge the Q-values. 
\lae{Concerning policy-gradient algorithms, \cite{MAPPO} suggest that Multi-Agent PPO (MAPPO) (cf. \S\ref{sect:MAPPO}) can also be a competitive baseline for cooperative multi-agent RL tasks.}
A parallel line of work is based on \textbf{coordination graphs}. In \cite{bohmer2020dgc}, a purely abstract graph represents each agent as a node, but the relations of each pairs are handled by \jac{MultiLayers Perceptrons (MLPs)}, which is restrictive. Instead, GNNs have been used in multi-agent RL to enhance the cooperation among agents, leveraging the graph whose nodes represent agents’ information. In \cite{jiang2018graphconv_rl} a GCN allows to expand the number of neighbors included in the calculation. DICG \cite{li2021dicg} is very similar, except that the structure of the graph is dynamically calculated with attention mechanisms. MAGE-X \cite{yang2023magex} (detailed in \S\ref{sect:magex:attn}) uses GCN paired with Gumbel Softmax \cite{jang2017gumbel_soft1,bhaskar2021gumbel_soft2} instead of attention to produce a dynamic discrete graph, more realistic for communication than attention.


\section{Contribution}
\lae{We first briefly recall relevant background on specific neural networks used in our model and on two related works at the foundations of our contribution. Then our MultiSoc model is presented in detail as well as the applied RL approach. In the following, agents refer to the robots and entity refers to robot or humans. }

 \subsection{Preliminary}

\subsubsection{Graph Neural Network}

Graph neural network (GNN) \cite{Scarselli:2009ku} is a neural network designed specifically to process and analyze data structured as a graph. Different architectures have emerged, of which the most well-known are GCN \cite{Kipf2017GCN} as convolutional methods and GAT \cite{velickovic2018gat} as spatial methods.
Both can be formalized similarly by aggregating information between neighboring nodes. The only difference is the calculation of the coefficients used in the aggregation process. Coefficients are static and directly dependent of the structure of the graph in GCN, while they are calculated dynamically with attention mechanism between neighboring nodes in GAT. This allows more flexibility but at the cost of greater calculation.

\subsubsection{Gumbel Softmax Transformer}
Gumbel Softmax Transformer (GST) \lae{\cite{huang2022gst}} is a neural network used for pedestrian motion prediction. Composed of 3 consecutive parts, 
only the first one, the \textbf{Edge Gumbel Selector}, will draw our attention. It computes a dynamical, complete and directed graph of the interactions between pedestrians, with their positions as nodes. A multi-head attention (MHA) produces attention between each possible edges and offers the possibility to connect each nodes with at most \(N_{head}\) other nodes, where \(N_{head}\) is the number of heads of MHA. The gumbel softmax trick allows to perform discrete clustering compatible with gradient descent. By choosing a low number of heads, we can then produce a sparse graph and vice versa. Tests on different types of crowds show significant differences depending on the number of heads and the density of the crowd.

\subsubsection{MAGE-X/AttnGraph}\label{sect:magex:attn}

\begin{figure}[htb]
    \centering
    \includegraphics[width=0.8\linewidth]{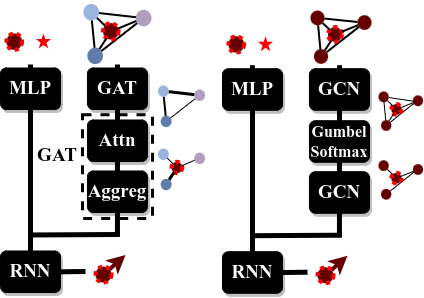}
    \caption{Oversimplified architectures of AttnGraph \cite{liu2023attngraph} and MAGE-X \cite{yang2023magex}. Agents (robots) are in red and agent of interest is surrounded by dotted line. (left) AttnGraph : At the end of each bloc is represented the graph actually computed and attention (edges width). (right) MAGE-X : At the end of each bloc is represented the graph actually computed.}
    \label{fig:comparaison}
\end{figure}

\lae{We now detail two GNN-based models 
to better highlight their commonalities and differences, summarized in Fig. \ref{fig:comparaison}.}   

\textbf{AttnGraph} \cite{liu2023attngraph} \lae{focuses on mono-robot social navigation}. It first uses a trajectory predictor (e.g. constant speed trajectory or more sophisticated ones as GST \cite{huang2022gst}) to dissipate uncertainty in the process. Then a complete graph  \lastmodif{is constructed, focusing only on the humans and ignoring the robot: each node is composed of the consecutive future positions of each human and each edge represents the visibility between humans}. This graph is submitted to an attention module (first GAT). 
This produces a new graph, with structure defined by attention and features influenced by each neighbors (middle graph in Fig. \ref{fig:comparaison} (Left)). Then, the robot is integrated in the graph submitted to an other GAT, presented in 2 blocks in Fig.\ref{fig:comparaison} (Attn + Aggreg) to highlight the parallelism with MAGE-X. 

Thus, the first GAT focuses on the analysis of the interactions between humans. The second GAT focuses on the robot and its links to the crowd (second graph with star structure centered on the robot in Fig. \ref{fig:comparaison} (left)).
Finally, the node of the robot is extracted and passes through a RNN, producing the action.
\newline

\textbf{MAGE-X} \cite{yang2023magex} focuses on multi-agent navigation problems. First agents goals are chosen in a centralized way before the navigation. This initial centralized assignment reduces the difficulty of predicting other trajectories, as their behaviors are influenced right the beginning. Moreover, it allows to transform the multi-agent navigation problem to multiple single-agent navigation tasks, in which each agent is required to reach the designated goal while avoiding collisions with others. A comparison can here be made with AttnGraph, in which the only agent (robot) must navigate to its goal while avoiding collisions with others (humans).

During navigation, MAGE-X is decentralized. A complete graph with each agent positions as nodes is submitted to a first GCN (cf. Fig. \ref{fig:comparaison} (Right)). Combined with a gumbel softmax, a discrete clustering on the edges is realized to  produce a sparser graph keeping only the most important neighbors of the agent of interest.

We can note here that the edge selection operation is similar (but not identical) to the one performed by the Edge Gumbel Selector of GST. Then, a second GCN analyses the sparser graph and the node of interest is extracted, concatenated with the intrinsic parameters of the agent, to obtain robot action through a RNN.

Thus, as highlighted by Fig.\ref{fig:comparaison}, the philosophy of these algorithms are close. Both used 2 GNNs and retract more and more the process on the agent/node of interest (encoded by the architecture in AttnGraph and with GCN combined with gumbel softmax for MAGE-X). 

However, the nature of the entities made the differences unavoidable. AttnGraph analyses humans and agent sequentially, while MAGE-X can analyse every entities from the beginning because they have the same nature.

\subsection{Our model}

 \begin{figure*}[htp]
     \centering
     \includegraphics[width=1.0\linewidth]{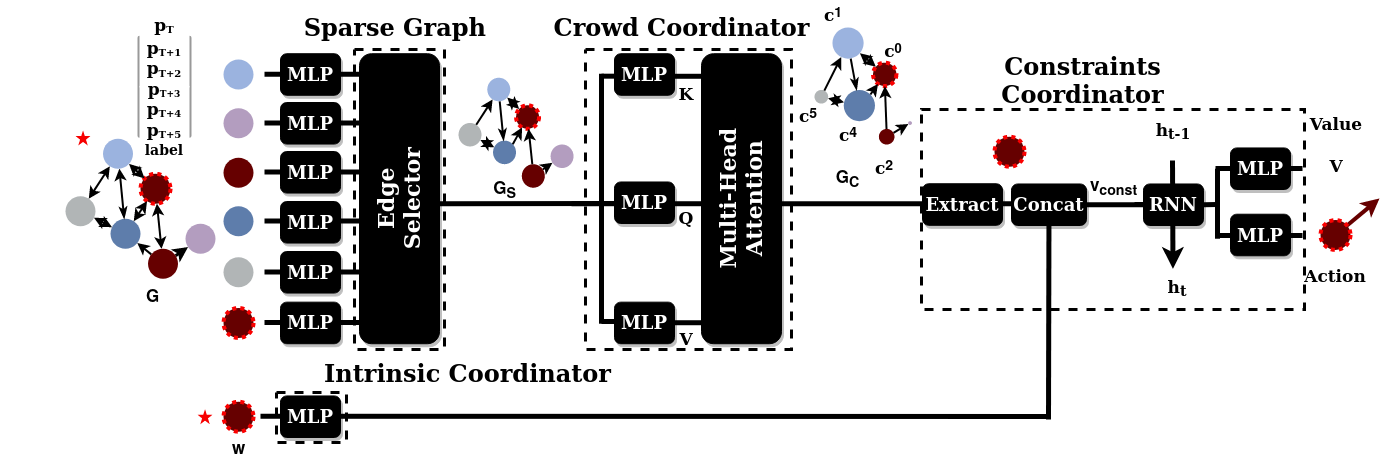}
     \caption{Overview of the MultiSoc architecture. \lae{For the agent of interest (surrounded by dotted line), the input is its intrinsic information and a graph limited to its FoV. Each node of the graph is composed of the current and consecutive predicted positions of the observed entities,} and by a label discriminating entities following their nature.}
     \label{fig:multisocv2}
 \end{figure*}

We present here our main contribution, MultiSoc, that is a model for learning multi-agent navigation among humans. MultiSoc can be seen as an homotopy between AttnGraph and MAGE-X. From the former, we keep the attention mechanism and the graph with predicted future positions of the entities. From the latter, we improve edge-selection and take up the graph merging early all the entities. 

\lastmodif{Moreover, unlike AttnGraph where entities ignore each other except for the only robot's consideration of humans, in multi-agent scenarios, interactions and visibilities among controllable agents are crucial for their coordination. That's why in MultiSoc, visibility among entities is critical, and the computation graph is based on this element.}
Indeed the algorithm has to merge both controllable and uncontrollable entities, all of them interacting with each other.
MultiSoc workflow for each agent $j$ is the following (cf. Fig. \ref{fig:multisocv2}):
\begin{itemize}
    \item[-] The Edge-Selector applies attention on nodes of a graph composed of predicted positions of each entities in the FoV of agent $j$. This produces a sparse directed graph with adjustable density given $N_{head}$. 
    \item[-] The Crowd Coordinator, a GAT with one layer of attention, is applied on the \lae{sparse} graph \lae{to compute node features influenced by neighbors}. Meanwhile, the Intrinsic Coordinator produces a broader summary of the constraint applied on the robot (constraint of goal).
    \item[-] Once the external constraints (Crowd Coordinator) have been correctly represented on the node representing the agent $j$, this node is extracted and concatenated with the constraint of goal (Intrinsic Coordinator).
    \item[-] Then a GRU, followed by two MLPs, produces both value and action, with respect to the previous hidden state and the information previously computed.
\end{itemize}

From a technical perspective, the GNNs included in the architecture allows:
 (i) A flexible computation taking into account as many entities as wanted. It is worth noting that humans and agents are included in the same graph (and not in the architecture itself as does AttnGraph) 
  (ii) A pseudo centralization in the decision acts on each agent in the graph (as node). By extension of the homogeneous paradigm, all other agents with enough information on their own observations, can be accurately understood by the concerned agent.
  (iii) An extendable receptive field increases the observation space from which the agent takes its action. As in \cite{jiang2018graphconv_rl}, each GNN layer extends the information perceived and allows nodes to be connected indirectly to more nodes.
\\
\subsubsection{Input data}
\paragraph{Trajectory prediction}
As in AttnGraph, the input information for each entity is a prediction of the trajectory. But instead of leaving the choice of the prediction method to the user, we systematically use the "constant speed" prediction. Because humans are highly unpredictable on the long term, a short-term prediction is enough. 
More importantly, AttnGraph results \cite{liu2023attngraph} do not provide great superiority of more complex prediction method for 5 timesteps trajectory. The constant speed method approximates the current speed of an entity by subtracting its current and previous positions. From that, the trajectory at constant speed and for 5 time-steps is calculated for each entity observed by the agent of interest.

\paragraph{Input Graph}
For the rest of the section, \(N\) will be the total number of entities in the simulation (a mask being applied on entities to represent the partial observation).
\newline
We obtain, from the trajectory predictor, the approximated future trajectory \(T_i = [p_i^t, ... , p_i^{t+5}]\) for each entity \(i\) with \(p_i^t\) the position of \(i\) at timestep \(t\). Along trajectories, we obtain the intrinsic information \(w^j = [p, v, g, \theta, r]\) of agent \(j\) with \(p\) its current position, \(v\) its speed, \(g\) its goal, \(\theta\) its angular heading and \(r\) its radius. 
Each agent has a limited FoV (depth and angle) and has access to the visibility matrix \(\matr{M^j} \in R^{N \times N}\) between each entity in its vision, defined by:
\begin{equation}
  M^j_{i,k} =
    \begin{cases}
      1 & \text{if \(i\) sees \(k\) and \(i\) is seen by the agent of interest $j$}\\
      0 & \text{otherwise}
    \end{cases}       
\end{equation}
\newline
\lae{For each agent \(j\), a graph \(G^j=(\matr{E},\matr{M}^j)\) is then formed where $\matr{E}$ is the node set with features $e_i$ of each node \(i \in [1,N]\) equal to \(T_i\) and a label discriminating entities following their nature (e.g. $label\in \{robot,human\}$). \(\matr{M^j}\) is the adjacency matrix for $j$.}
The state \(s_t^j = [w^j, G^j]\) forms the input given to the MultiSoc model, \lae{as illustrated in Fig. \ref{fig:overview} (Bottom) with a graph limited to each agent FoV}. 
\subsubsection{Edge-Selector}

\begin{figure}[tb]
    \centering
    \includegraphics[width=1.\linewidth]{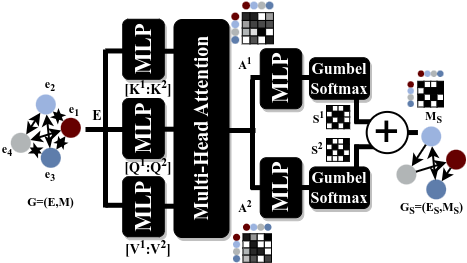}
    \caption{\lae{Overview of the Edge-Selector architecture producing a sparse graph $G_S$ with a MHA module with 2 heads.}}
    \label{fig:edge_selector}
\end{figure}

The first component of MultiSoc is an Edge-Selector detailed in Fig. \ref{fig:edge_selector}. \lae{It applies attention on nodes of $G$ to produce a sparse directed graph keeping only the most interesting interactions between entities}. Edge-Selector is a tractable adjustment of the Edge Gumbel Selector from GST \cite{huang2022gst}, 
where the attention is applied on the set of possible edges (\(N\times N\) edges with \(N\) the number of entities). For a supervised algorithm, this computation is not a limit but for RL, which learns online, the speed of the algorithm is crucial. Thus, we designed Edge-Selector based on the attention between nodes only. The remaining mechanisms are similar to the ones found in GST.
\newline
First, a MHA with $N_{head}$ heads is applied between every nodes $i$ \lae{(cf. Fig. \ref{fig:edge_selector} (Left)). For each head $k$, the input matrix $\matr{E}$ with features $e_i$ of all nodes $i \in [1,N]$ is linearly transformed using learned weight matrices: $\matr{Q}= \matr{E}\matr{W}_Q, \matr{K}=\matr{E}\matr{W}_K, \matr{V}=\matr{E}\matr{W}_V$. Then, $\matr{Q}, \matr{K}, \matr{V} $ are divided along the features axis into $N_{head}$ parts such that $\matr{Q} = [\matr{Q}^1 .. \matr{Q}^{N_{head}}], \matr{K} = [\matr{K}^1 .. \matr{K}^{N_{head}}], \matr{V} = [\matr{V}^1 .. \matr{V}^{N_{head}}]$. The scaled dot-product attention is computed as follows for each head:
\begin{equation}
\matr{A}^k = Softmax(\frac{\matr{Q}^k (\matr{K}^k)^T}{\sqrt{d_n}})\matr{V}^k
\end{equation}
where $d_n$ is the dimension of the nodes. Thus MHA applied to all node features $e_i$ of the graph $G$ produces attention score \(a_{i,j}^k\) between agents \(i\) and \(j\) for the \(k^{th}\) head.  }

As for GST, the mask $\matr{M_S}$ for the edges is calculated with Gumbel-Softmax  \lae{(cf. Fig. \ref{fig:edge_selector} (Right))}:
\begin{equation}
    \begin{gathered}
s_{i,j}^k = Softmax_{j} \frac{MLP(a_{i,j}^k) + g}{\tau} \\
m_{i,j} = \frac{1}{N_{head}} \sum_{k=0}^{N_{head}} s_{i,j}^k
\end{gathered}
\end{equation}
where \(N_{head}\) is the number of heads, \(g\) is sampled from the Gumbel distribution, \(MLP\) is a linear layer, \(\tau\) is the temperature parameter used in Gumbel-Softmax trick and \(m_{i,j}\) \lae{are the coefficients of the adjacency matrix $\matr{M_S}$} produced as a mask on the edges. Thus, in the final sparse graph $G_S=(\matr{E}_S,\matr{M}_S)$:
\begin{itemize}
     \item[-] the feature of each node in $\matr{E}_S$ is the attention scores for agent $i$.
     \item[-] there is an edge between the \(i^{th}\) and \(j^{th}\) nodes if \(m_{i,j} \ne 0 \).  Moreover, \textbf{each node $i$ has at most $N_{head}$ edges}.
\end{itemize}
\lae{The Edge-Selector realizes a discrete clustering on the edges and allows to introduce the parameter \(N_{head}\) as a constraint of density on the dynamical graph. Greater \(N_{head}\) is, denser the graph will be.}

\subsubsection{Crowd Coordinator}

Now that a \lae{sparse graph with entities in the FoV of the agent of interest has been calculated, we can propagate information between neighboring nodes to compute node features influenced by their neighbors.}  

\lae{Various approaches are used in the literature for this.} DICG \cite{li2021dicg} , like MAGE-X \cite{yang2023magex}, applied a GCN on the graph while AttnGraph \cite{liu2023attngraph} used a \lae{simple  attention mechanism analogous to a GAT on a star graph. It has to be noted that} an ablation study on MAGE-X proved that GCN could be replaced by attention mechanism without dramatic loss in accuracy (7\% less success with this replacement \cite{yang2023magex}).
\lae{We decide to use a GAT \cite{velickovic2018gat} as attention mechanism to analyse the sparse graph. This choice applies naturally on the directed sparse graph and allows us to remain consistent with AttnGraph, which we aim to expand for multi-agent social navigation.}
\newline
\lae{Considering the sparse graph $G_S$ produced by the Edge-Selector} (cf. Fig. \ref{fig:edge_selector}), with \(h_{i}\) being the features of the \(i^{th}\) node and \(\matr{H}\) the matrix concatenating all \(h_i\), the attention is defined by:
\begin{equation}
Attention(\matr{H}) = Softmax(\frac{\matr{Q} \matr{K}^T}{\sqrt{d_k}})
\end{equation}
where \(\matr{Q} = \matr{H}\matr{W}_Q\), \(\matr{K} = \matr{H}\matr{W}_K\), \(\matr{W}_Q\) and \(\matr{W}_K\) being learned weight matrices, \(d_k\) is the dimension of the node and
\begin{equation}
Softmax_j(d_{i,j}) = \frac{e^{d_{i,j}}}{\sum_{k \in N_i} e^{d_{i,k}}}
\end{equation}
where \(d_{i,j}\) is the \((i,j)\) coefficient of \(\frac{Q K^T}{\sqrt{d_k}}\) and \(N_i\) is the neighborhood of the node \(i\) in the sparse graph (deduced from $\matr{M_S}$).
\newline
Each attention head $k$ produces \(\alpha_{i,j}^k = Attention^k(\matr{H})_{i,j}\). Finally, the nodes are weighted summed together according to their neighborhood in the sparse graph. The crowd coordinator then produces a graph $G_C$ with features \(c^i\) for each node $i$ (cf. Fig. \ref{fig:multisocv2}):
\begin{equation}
c^i = Concat_k(\sum_{j \in N_i} \alpha_{i,j}^k h_j \matr{W}_V)
\end{equation}
where \(Concat_k\) is the function concatenating all the vectors over the heads (and then merges the vectors produced by each head in an unique vector) and \(\matr{W}_V\) is a learned weight matrix. It is possible to add an activation function as \(\sigma = ReLu\) before concatenating over the heads and more importantly, it is also possible to add new layers. We do not add these supplements because the graphs are small enough in our problem (small field of view producing relatively small graphs).  

\subsubsection{Constraints Coordinator}
\paragraph{Extract and Concat}
The crowd coordinator produces a graph representing the influence of each node on the others. In line with this idea, only the node of interest has to be kept. We can observe that, when we use a GAT with one layer, as we do, only the neighborhood of the node of interest has an influence on it. We can then keep only the edges between the agent and its neighborhood in the previous GAT, as it is done in AttnGraph. However, to be complete, we keep the general formulation and do not truncate any edges before the operation, keeping so possible generalisation.
\newline
Thus, we extract \lae{from $G_C$} the node representing the agent of interest and concatenate it with the output of the intrinsic coordinator, which is simply an MLP. By doing so, we obtain both the constraints of \lae{the other entities on the agent of interest} and the constraint of reaching the goal in one vector \(v_{const}\).

\paragraph{RNN}
We use a gated recurrent unit (GRU) to keep consistency with the previous action:
$h^t = GRU(h^{t-1}, v_{const})$
where \(h^{t-1}\) is the hidden state produced at last timestep and \(v_{const}\) is the vector of constraints. The value and the action for RL are then produced by passing \(h^t\) through 2 different MLPs.

\subsection{Reinforcement Learning}
\lae{We model the scenario as a Multi-Agent Markov Decision Process}. MultiSoc follows CTDE paradigm. The centralization part lies in the fact that all the agents are trained with the same MultiSoc model. \lae{But execution is decentralized as at each time-step, each agent passes through the MultiSoc model its intrinsic information and positions of entities in its FoV. This produces the action the agent will take. Thus, each agent receives a reward and the simulation transits to a next state according to an unknown state transition, taking into account humans and other agents actions.}

\subsubsection{MAPPO}\label{sect:MAPPO}
Concerning the RL algorithm, AttnGraph \cite{liu2023attngraph} uses Proximal Policy Optimization (PPO) \cite{PPO}, a model-free policy gradient algorithm widely used.

The most direct heir of PPO in multi-agent paradigm is Multi-Agent PPO (MAPPO) \cite{MAPPO} used by MAGE-X \cite{yang2023magex}. Even if adaptations to the multi-agent are observed in MAPPO (e.g. value normalization), it differs mainly from PPO by the parameters fine-tuning. Indeed, in multi-agent, it is observed \cite{MAPPO} that neural network training is really sensitive to the number of epochs. To stay aligned with AttnGraph and MAGE-X, we also use MAPPO.
The parameters will be close to the ones used by AttnGraph, the main difference will remain in the number of epochs, multi-agent being empirically better trained with fewer epochs.

\subsubsection{Reward}
We adopt the reward function used in AttnGraph \cite{liu2023attngraph} with some modifications to consider multi-agents. The penalty for being in the predicted path of an entity (human or agent) is: 

\begin{equation}
    \begin{gathered}
    r^{j,i}_{pred}(s_{t}) = min_{k \in [1,5]}((\mathbb{1}^{t+k}_j,i) \frac{r_c}{2^k}) \\
    r^j_{pred}(s_t) = min_{i \in [1,N]} (r^{j,i}_{pred}(s_{t}))
    \end{gathered}\label{eq:penalty}
\end{equation}
\newline
where \(s_t\) is the state of the agent of interest $j$, \(r_c\) the penalty for collision and \((\mathbb{1}^{t+k}_j,i)\) indicates whether the agent \(j\) collided with the \(k^{th}\) predicted position of the entity \(i\).
\newline
The potential based reward guides the agent \(j\) to approach the goal: \(r^j_{pot} = -d^{j,t}_{goal} + d^{j,t-1}_{goal}\) with \(d^{j,t}_{goal}\) the distance of \(j\) to its goal at timestep $t$.
\newline
The complete reward for agent \(j\) \lae{doing action $a_t$ in state $s_t$} is then:

\begin{equation}
  R^j(s_{t},a_{t}) =
    \begin{cases}
      r_c & \text{if j collides any entity}\\
      r^j_{pot} + r^j_{pred} & \text{otherwise}
    \end{cases}       
\end{equation}

We emphasize that there is no reward when reaching the goal. Indeed 
an episode is not completed until all agents have either collided or reached their goals.
Thus each agent that reaches its goal must wait the others to finish. If there was a reward for reaching a goal, one agent would wait the other with infinite reward.

\section{Experimentations}

\subsection{\lae{Simulation Environment}}
\paragraph{Simulator}
We extend the CrowdNav\footnote{\href{https://github.com/Shuijing725/CrowdNav_Prediction_AttnGraph}{https://github.com/Shuijing725/CrowdNav\_Prediction\_AttnGraph}} mono-agent simulator used in AttnGraph \cite{liu2023attngraph} for a multi-agent version. First, we improve the use of matrices especially concerning the visibility, which is central in our work. Second, we implemented our multi-agent simulator \textbf{MultiCrowdNav}\footnote{Code will be available online if the paper is accepted.} based on multi particle environments (MPE) \cite{lowe2017mpe}, in which small particle agents must navigate and communicate. It facilitates greatly the migration from PPO to MAPPO, as MAPPO already supports MPE environments.  %

\paragraph{Crowd Simulation}

\lae{The simulation of human interactions is essential for agent learning in a context of social navigation.} A complete model requires a huge amount of information to be taken into account, and becomes computationally intractable. Despite their lack of realism, methods such as ORCA \cite{ORCA} or social force (SF) \cite{SFM} are often used to simulate humans (for learning and testing). It's also worth noting, according to \cite{mavrogiannis_core_2023} (\S4.3.1),  that ORCA is mostly used for testing, while social force brings more adversity to the agent. The combination of the two would seem to be a prerequisite if we are to entertain the idea of a real-life implementation. \lae{Thus in our simulator each human can be controlled by ORCA or SF and some experiments will be done with heterogeneous human policies.}

\paragraph{Scenario} 
\lae{The scenarios are initialized with $H$ humans arranged on a circle with some noise on their positions. Human goals are chosen so that they must cross the circle to reach an opposite point. Humans react only to other humans but not to robots (adversarial crowd). A new random goal is assigned to a human as soon as it reaches its goal. 
At initialisation, $R$ agents are also laid out randomly with their own goals (positioned and assigned randomly\footnote{Unlike MAGE-X, agents can not exchange their goals with each other at the beginning of the episode. }). When an agent collides with another entity (human or agent) or reaches its goal, it keeps moving but no longer counts in the metric. Therefore, it is still considered as a moving obstacle for the remaining agents (considered in their penalty reward (cf. eq. \ref{eq:penalty})). An episode is over when all agents have either collided (collision) or reached their goal (success). For more details concerning the simulation (kinematics, action space, sensor range, ...) the reader can refer to \cite{liu2023attngraph}.}

\paragraph{Metrics} 
\lae{Our metrics include navigation and social metrics traditionally used in multi-robot and social robot  navigation.}
\lae{The \textbf{success rate} is the number of agents that reached their goals to the total number of agents, evaluated on all test episodes. }

Safety is mainly summarized by the \textbf{collision rate}, \lae{i.e. the number of agents colliding with other entity (humans or not).}  Once an agent has reached its goal, if at least one other agent has not yet reach its goal, collisions are no longer counted in the score for a succeeded agent. The proximity of the agent to humans can also be considered \lae{to evaluate the social awareness of the agents.} We used the \textbf{intrusion ratio} as the percentage of time the agent was "too close" to a human \lae{averaged over all episodes} ("too close" is defined as in \cite{liu2023attngraph} with a distance defining the space "close" to an individual).
To compare relative performance between algorithms without comparing to a theoretical optimal solution that is not available, several criteria are defined. \textbf{Travel time} and \textbf{travel length} are mean duration, resp. length, of the trajectories between the initial position to the goal (or ending position if not reached) for all agents. The \textbf{reward} is the mean reward obtained \lae{by each agent} at the end of episode.

\subsection{Results}

We now present and analyze results obtained with MultiSoc in a set of various scenarios (number of agents/humans) and compare it with baseline (AttnGraph). Models are trained during $N_{train}$ timesteps. Training hyperparameters can be found in the Appendix. Evaluations are conducted on $N_{test}$ random unseen episodes, each consisting of 150 timesteps. 
Videos and simulation screenshots showing the behavior learned by the agents with MultiSoc in different scenarios are available in supplementary materials.

\subsubsection{Baseline comparison}

\begin{table}[!t]
\caption{Baseline comparison. \lae{$N_{train}=20M$ of timesteps and $N_{test}=1000$ episodes (150 timesteps) with $seed=1000$.} 
During training and test ORCA policy is used for humans navigation. \lae{Reward metric is not given for AttnGraph because it is not comparable to MultiSoc (AttnGraph uses a reward for reaching the goal)}.}
\label{tab:baslineCompare}

\setlength{\tabcolsep}{2.85 pt}
\setlength{\extrarowheight}{5pt}
\begin{center}
\begin{tabular}{
 l |
  *{6}{l}|
  *{2}{l}|
  *{2}{l}
}
\toprule
  
  Models & \rotatebox{90}{\shortstack[c]{Success \\ $\rightarrow$}} & \rotatebox{90}{\shortstack[c]{Collision \\ $\leftarrow$}} & \rotatebox{90}{\shortstack[c]{Intrusion \\  $\leftarrow$ Ratio}} & \rotatebox{90}{\shortstack[c]{Travel\\ $\leftarrow$ Time}} & \rotatebox{90}{\shortstack[c]{Travel \\ $\leftarrow$ Length}} & \rotatebox{90}{\shortstack[c]{Reward\\ $\rightarrow$}} & \rotatebox{90}{R $\; \;$   Train} & \rotatebox{90}{H} & \rotatebox{90}{R $\; \;$  Tests} & \rotatebox{90}{H} \\
\hline
  Attngraph\footnotemark[4]  & $0.92$ & $0.05$  & $6.51$ & $15.47$ & \textbf{13.99}  & - & \multirow{2}{*}{1} & \multirow{2}{*}{20}  & \multirow{2}{*}{1} & \multirow{2}{*}{20}  \\
    MultiSoc  & \textbf{0.96} & \textbf{0.03}  & \textbf{4.33}  & \textbf{14.70}  & ${14.79}$ & \textbf{19.10}  & &  & & \\
\hline
    Attngraph\footnotemark[4]  & 0.85 & 0.13  & 5.84 & 15.81 & \textbf{14.38}  & - & 1 & 20  & \multirow{2}{*}{5} & \multirow{2}{*}{20} \\
    MultiSoc\footnotemark[5]  & \textbf{0.94} & \textbf{0.04}  & \textbf{3.42}  & \textbf{15.35}  & 15.87 & 17.24 & 1 & 20 & & \\
\hline
    AttnGraph\footnotemark[4]   & 0.68 & 0.31  & 11.92  & 16.39  & 14.49 & - & 1 & 20 & \multirow{3}{*}{6} & \multirow{3}{*}{6} \\
    MultiSoc  & 0.81 & \textbf{0.02}  & \textbf{3.98}  & 16.62  & 20.02 & \textbf{5.91} & 1 & 20 &  &  \\
    MultiSoc  & \textbf{0.85} & 0.14  & 13.61  & \textbf{12.8}  & \textbf{12.28} & -1.85 & 5 & 20 &  &  \\
\hline
\end{tabular}

\end{center}
\end{table}


\begin{table*}[hbt!]
\caption{ 
MultiSoc model qualification. Grey cells are variables variability, better results are in bold, $R$ and $H$ stand for number of Robots and Humans, $H-Policy$ is the human navigation policy \lae{during tests (ORCA is still used during training). Tests are done on $N_{test}$ episodes of 150 timesteps with $seed=1000$.($60k$ test timesteps means $N_{test}=400$ and $150k$ means $N_{test}=1000$).} 
}
\label{tab:results}
\begin{center}
\begin{tabular}{
  l| 
  *{6}{l} |
  *{4}{l}|
  *{4}{l}
}
\toprule
\multicolumn{1}{c|}{} &
  \multicolumn{6}{c|}{Metrics} &
  \multicolumn{4}{c|}{Training Conditions} &
  \multicolumn{4}{c}{Test Conditions} \\
 & \rotatebox{90}{\shortstack[c]{Success \\ $\rightarrow$}} & \rotatebox{90}{\shortstack[c]{Collision \\ $\leftarrow$}} & \rotatebox{90}{\shortstack[c]{Intrusion \\  $\leftarrow$ Ratio}} & \rotatebox{90}{\shortstack[c]{Travel  \\ $\leftarrow$ Time}} & \rotatebox{90}{\shortstack[c]{Travel \\ $\leftarrow$ Length}} & \rotatebox{90}{\shortstack[c]{Reward\\ $\rightarrow$}} & \rotatebox{90}{\shortstack[c]{N$_{head}$}} & \rotatebox{90}{\shortstack[c]{N$_{train}$\\ timesteps}} & R & H & \rotatebox{90}{\shortstack[c]{Test\\ timesteps}} & R & H & H-Policy \\

\hline
\multirow{6}{*}{\shortstack[C]{1. Pure Multi-agent \\ scalability }}
 & \textbf{0.91} & \textbf{0.03}  & \textbf{1.1}  & \textbf{12.27} & \textbf{13.81} & \textbf{18.75}  & \multirow{6}{*}{4} & \multirow{6}{*}{20M} & \cellcolor{ashgrey!25} 3  & 0 & \multirow{6}{*}{150k} & \cellcolor{ashgrey!25} 3  & 0 & \multirow{6}{*}{ORCA}\\
  & 0.81 & 0.12 & 4.06 & 15.17 & 16.86 & 8.55  &  &  & \cellcolor{ashgrey!25} 3  & 0 & & \cellcolor{ashgrey!25} 10 & 0 & \\
  & 0.59 & 0.07  & 7.26 & 17.78 & 19.46 & -4.54 &  &  & \cellcolor{ashgrey!25} 3  & 0 & & \cellcolor{ashgrey!25} 20 & 0 & \\\cline{2-7}
  & 0.92 & \textbf{0.00}  & 0.75 & \textbf{11.83} & \textbf{12.29} & \textbf{20.47}  &  &  & \cellcolor{ashgrey!25} 10 & 0 & & \cellcolor{ashgrey!25} 3  & 0 & \\
  & \textbf{0.94} & \textbf{0.00}  & \textbf{1.92} & 13.54 & 13.49 & 18.29  &  &  & \cellcolor{ashgrey!25} 10 & 0 & & \cellcolor{ashgrey!25} 10 & 0 & \\
  & 0.89 & 0.01  & 4.31 & 15.76 & 16.20 & 13.2 &  &  & \cellcolor{ashgrey!25} 10 & 0 & & \cellcolor{ashgrey!25} 20 & 0 & \\
\hline
\multirow{3}{*}{\shortstack[C]{2. Human Policy \\ Robustness}}
  & 0.92 & 0.07 & 7.60 & 13.59 & 13.18 & 14.96 & \multirow{3}{*}{4} & \multirow{3}{*}{20M} & \multirow{3}{*}{3} & \multirow{3}{*}{17} & \multirow{3}{*}{150k} & \multirow{3}{*}{3} & \multirow{3}{*}{17} & \cellcolor{ashgrey!25} ORCA\\
 & \textbf{0.94} & \textbf{0.06} & \textbf{7.59} & 13.45 & 13.12 & \textbf{15.35} & & & & & & & &\cellcolor{ashgrey!25} ORCA+SF\\
 & 0.66 & 0.33 & 8.27 & \textbf{12.32} & \textbf{10.66} & 5.47 & & & & & & & & \cellcolor{ashgrey!25} ORCA+FoV\\
 \hline
\multirow{4}{*}{\shortstack[C]{3. Scalability}}
  & \textbf{0.96} & \textbf{0.03} & \textbf{4.33} & 14.70 & 14.79 & \textbf{19.07}  & \multirow{4}{*}{4} & \multirow{4}{*}{20M} &\cellcolor{ashgrey!25} 1 & 20 & \multirow{4}{*}{150k} & \cellcolor{ashgrey!25} 1 & 20 & \multirow{4}{*}{ORCA}\\
 & 0.95 & \textbf{0.03} & 8.94 & 14.33 & 13.82 & 17.67  & & & \cellcolor{ashgrey!25} 5 & 20 & &\cellcolor{ashgrey!25} 1 & 20 & \\
 & 0.89 & 0.08 & 6.79 & \textbf{14.22} & \textbf{13.68} & 14.0  & & & \cellcolor{ashgrey!25} 5 & 20 & &\cellcolor{ashgrey!25} 5 & 20 & \\
 & 0.89 & 0.08 & 6.95 & 14.85 & 14.40 & 12.68 & & & \cellcolor{ashgrey!25} 5 & 20 & &\cellcolor{ashgrey!25} 10 & 20 & \\
 \hline
\multirow{3}{*}{\shortstack[C]{4. Density Factor}}
 & 0.88 & 0.05 & 3.39 & 18.10 & 18.93 & 15.78 & \cellcolor{ashgrey!25} 2 & \multirow{3}{*}{20M} & \multirow{3}{*}{5} & \multirow{3}{*}{15} & \multirow{3}{*}{60k} & \multirow{3}{*}{5} & \multirow{3}{*}{15} & \multirow{3}{*}{ORCA}\\
  & \textbf{0.91} & \textbf{0.02} & \textbf{2.39} & \textbf{17.36} & \textbf{17.30} & \textbf{17.95} & \cellcolor{ashgrey!25} 4 & & & & & & & \\
  & 0.86 & \textbf{0.02} & 2.83 & 18.38 & 17.86 & 17.46 & \cellcolor{ashgrey!25} 8 & & & & & & & \\
\hline
\end{tabular}

\end{center}

\end{table*}

\jac{We first compare our MultiSoc model to Attngraph \cite{liu2023attngraph} that, as far as we know, actually overcomes other deep RL models in mono-robot navigation in crowd environment.  }
\jac{First, the modifications made to the simulator and the paradigm shift from mono-agent to multi-agent RL brings about a preliminary improvement in terms of learning time. Training Attngraph on $20$ Millions of timesteps with CrowdNav simulator actually takes 40h (average) in our training condition ( \lastmodif{4 Xeon E5-2640v3 CPUs, with 32GB of memory and one NVIDIA GK210 GPU}), whereas training Attngraph with our new simulator MultiCrowdNav costs no more than 20h (average) in the same conditions. }

\jac{ Second, as shown in Table \ref{tab:baslineCompare}, our \textbf{MultiSoc model overcomes Attngraph especially when several robots are involved}.}

\jac{In a first test, both are compared during a learning and training phase with $1$ robot and $20$ humans}. \jac{The results} confirm the similarity between MultiSoc and AttnGraph \lae{in mono-robot conditions}, as they are similar under the same conditions. The better performance of MultiSoc does not allow us to decide on any superiority. AttnGraph is very hard to train, while MultiSoc suffers from the same single-agent limitations. Nevertheless, the two models remain close in their results. 

\footnotetext[4]{\lastmodif{The Attngraph used is a pre-trained version of the model provided by authors.}}
\footnotetext[5]{ This model was trained over 30 Millions of timesteps.}
\footnotetext[6]{More results are given in the Appendix.}

\jac{Several robots are introduced during test phase only in a second experiment ($5$ robots and $20$ humans). Each robot executes individual trained Attngraph \lae{with other robots considered as humans}. In such condition, the success rate of Attngraph decreases to $0.85\%$ showing the model's difficulty in handling a heterogeneous crowd mixed of robots and humans. 
Despite the fact that our MultiSoc was trained with only one robot, its still remains resilient ($0.94\%$ of success) when several robots are involved. This result shows \textbf{a better generalization of our model when dealing with a crowd composed of robots and humans}.}

\jac{Finally, in a third test, a crowd composed of $6$ robots and $6$ humans is used during testing. In such condition, the performance of Attngraph drops down to $0.68\%$ of success. This is mainly due to the difficulty of Attngraph to predict behavior of other robots. This leads to "panic" situations and weird behavior when a robot encounters other robots. 
Regarding MultiSoc, coordinating multiple robots remains challenging \lae{in a more balanced crowd}, but it maintains a high success rate whether it learns with $1$ or $5$ robots.
}
\erw{More importantly, MultiSoc can properly transfer mono-agent learning to multi-agent problem, meaning that it can use its own behavior to predict similar entities, thanks to the labels on the node.}

\subsubsection{Multi-agent without humans}

In Table \ref{tab:results}.1, Multi-Soc has been trained on respectively 3 and 10 robots \textbf{without humans}, arranging agents and their goals randomly on the map. 
The multi-agent's great adaptability is particularly noteworthy for tests with 10 robots when model was trained with 3 robots (performance of 81\% success rate for 12\% collision rate). This is even more striking for a model trained with 10 robots and tested with 20 (low performance loss).
We can hypothesise that this \textbf{scalability} is due to the management of trajectories by the neural networks. As observed in \cite{mavrogiannis_core_2023}, deep-learning-based navigation methods suffer from short-range. This allows the presented network to focus only on immediate trajectories (i.e. fewer configurations than for longer-term trajectories). As a result, transfer is more efficient, since it's the immediate environment that impacts the strategies and not the overall number of agents, at the cost of a loss of "centrality".

\subsubsection{Social (Multi-agent with humans)}

The results presented in Table \ref{tab:results}.2 \lae{illustrate \textbf{heterogeneous human policies management of MultiSoc\footnotemark[6]}, which is a flaw of current works on social robot navigation according to \cite{mavrogiannis_core_2023} (\S4.3.1)}. MultiSoc is trained with several agents, several humans and tested under various social conditions.\textbf{ The architecture handles the mix of human policies very well} \lae{even though training was done with homogeneous human policies (ORCA).} Indeed tests mixing ORCA and Social Force (ORCA+SF) do not bring any additional difficulties to the model \jac{($94\%$ of success rate)}. However, having a reduced field of view for humans (ORCA+FoV) remains a clear limitation. In fact this \jac{situation} creates a very complex crowd, far removed from the crowd on which the models were trained \jac{ and brings more unpredictable human behaviors}.

The results presented in Table \ref{tab:results}.3 \lae{further demonstrate the \textbf{scalability capacities of MultiSoc}}. Trainings with different numbers of robots delivered stable performance, but it's worth noting that a given training can also adapt to situations with more robots. \jac{The results demonstrate} that training with 5 robots achieves very good results in single-robot (95\% success rate), but also good results with 10 robots (89\% success rate). \textbf{The scalability observed in the pure multi-agent case transfers well to the social case.}
\newline
These results also open up new possibilities for the single agent case. Indeed, AttnGraph suffered from \jac{the difficulty of reproducing the model's learning process} (according to the authors of AttnGraph and us). MultiSoc suffered from the same issue in single-agent mode, but multi-agent training is empirically more stable and faster. \lastmodif{Models converge faster at the cost of a slightly longer computation time,} since the MultiSoc model is fed N times more than in single-agent (where N is the number of agents) and with more "balanced" data concerning positions, since the agents are far from each other. This results in more stable, less biased, faster training for robust single-agent transfers. 

\jac{Finally,} the results in Table \ref{tab:results}.4 illustrate the \textbf{usefulness of the density factor }(\(N_{head}\)) retrieved from the Edge Selector of GST \cite{huang2022gst}. 
As noted previously, AttnGraph has not proven the benefit of its use of GST. We hypothesize that the complete graph used throughout AttnGraph lost the essential information carried by GST, that is the human link management. By extracting only a short-term forecast, \jac{we assume that} AttnGraph lost the deeper analysis, leaving only a too-compact summary. Using a simplified version of the Edge Selector allows us to better integrate it into the architecture.
\textbf{Varying the density factor \(N_{head}\) allows to obtain different results for the same crowd}. The model performs best for \(N_{head}\)=4 (0.91\% success rate) for the tested scenario, which is fairly consistent with our observations of the crowd, since agents will most often see fewer than 5 humans in their vicinity.

\section{Conclusion}
In this paper we propose MultiSoc, a new method for learning multi-robot socially aware navigation strategies. 
MultiSoc leverages graph-based representations combined with attention mechanisms to capture heterogeneous interactions in the crowd and various influences of each entity on the others. Especially the Edge Selector we integrate in our model allows to take into account the crowd density. The experiments show that our MultiSoc outperforms the main baseline and deals with heterogeneous human policies. It also demonstrate MultiSoc scalability capacities and the usefulness of the density factor we introduced.

\lastmodif{
Future work needs to investigate how our model deals with a complex environments including obstacles such as sparse environment, corridor, room, ... Moreover, interacting with a crowd with different behaviors (e.g cooperative, adversarial, social groups) may affect the performance of our model and should be studied. Finally, testing our social navigation model on real robot fleet in a real crowd situation remains essential to validate our model.
}

\textbf{Acknowledgement} ---
This work has been partially funded by ANR project DeLiCio (ANR-19-CE23-0006-DA). 

\bibliographystyle{IEEEtran}
\bibliography{sample}

\section*{Appendix}

\subsection*{Training hyperparameters } 
In Table \ref{tab:mappohyperparam} are given the common training hyperparameters we used for MAPPO \cite{MAPPO}. Other hyperparameters specific to MultiSoc can be found in Table \ref{tab:otherhyperparam}. The detailed architecture of MultiSoc is shown in Table \ref{tab:multisocnn}.

\subsection*{MultiCrowdNav simulator} 
\begin{figure*}[bt]
     \centering
     \includegraphics[width=1.0\linewidth]{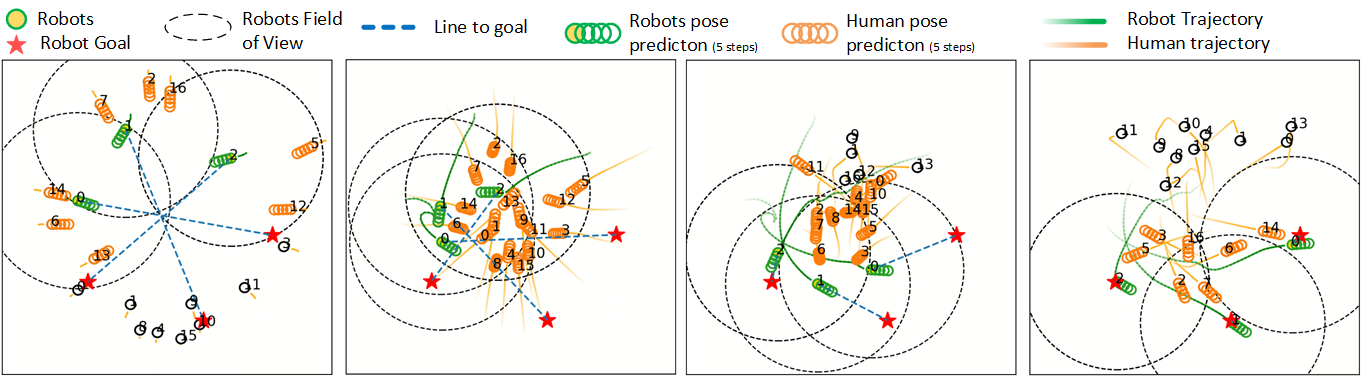}
     \caption{Screenshots of a scenario (in chronological order from left to right) with 3 robots traveling among 20 humans in the MultiCrowdNav Simulator ($N_{head}=4$). 
     }
     \label{fig:simu}
 \end{figure*}

In Figure \ref{fig:simu}, screenshots of our simulator (in chronological order) are given for a scenario with 3 robots traveling among 20 humans. From left to right :
\begin{enumerate}
    \item[1-] The initial situation with robots in green and their field of view (FoV) in a dashed circle is shown. The dashed blue line connects each robot to its goal (red star). Humans in the FoV of robots are in orange; others are in black. 
    Predicted poses on 5 time steps are represented with 5 circles for each entity in the FoV of a robot. The trajectories followed by each entity are given with a solid line that fades over time.
    \item[2-3-] Given that human goals are chosen so that they must cross the circle to reach an opposite point, we can observe here that all humans have converged in the center of the scene. In order to reach their goals, the robots will also have to traverse the central space by default, leading all entities to cross paths in the center of the environment. The robots will then have to manage challenging crowd navigation and implicit coordination with other entities to avoid collisions and intrusion in human safe space. We can observe that robots 0 and 1 decided to navigate around the high-density area. Especially robot 0 chose to navigate around on the side where it predicts humans won't be. As for robot 2, it traverses the crowd while locally managing interactions.
     \item[4-] All three robots have successfully reached their objectives without collisions.
\end{enumerate}

\begin{table}[tb]
\begin{center}
\begin{tabular}{ c|c }
\hline
{common hyperparameters} & value \\
\hline
nrolloutthread & 16 \\
numminibatch & 2 \\
episode length & 50 \\
data chunck length & 50 \\
num env steps& 20 000 000 \\
ppo epoch &5 \\
gain& 0 \\
lr & 4e-5 \\
critic\_lr&  4e-5 \\\hline

\end{tabular}
\caption{\label{tab:mappohyperparam} MAPPO Hyperparameters }
\end{center}
\end{table}

\begin{table}[tb]
\begin{center}
\begin{tabular}{ l|c }
\hline
{architecture hyperparameters} & value \\
\hline
human\_node\_rnn\_size & 128 \\
human\_node\_output\_size & 256 \\
edge\_selector\_embedding\_size & 32 \\
agent\_embedding\_size & 64 \\
human\_node\_embedding\_size & 64 \\
human\_human\_edge\_embedding\_size & 32 \\
attention\_size & 64 \\
human\_node\_input\_size & 3 \\
human\_human\_edge\_input\_size & 2 \\
human\_human\_edge\_rnn\_size & 256 \\
edge\_selector\_emb\_size & 512 \\
edge\_selector\_num\_head & 4 \\
mha\_emb\_size & 256 \\
mha\_num\_head & 8 \\
\hline

\end{tabular}
\caption{\label{tab:multisocnn} MultiSoc architecture Hyperparameters }
\end{center}
\end{table}

\begin{table}[tb]
\begin{center}
\begin{tabular}{ c|c }
\hline
{Other hyperparameters} & value \\
\hline
temperature at beginning & 5 \\
base temperature & 0.05 \\
min temperature & 0.03 \\
collision penalty $r_c$ & -20 \\
\hline

\end{tabular}
\caption{\label{tab:otherhyperparam} Other Hyperparameters }
\end{center}
\end{table}

\subsection*{Additional Results} 

In Table \ref{tab:results} are given additional results concerning \textbf{the heterogeneous human policies management
of MultiSoc}.


\begin{table*}[hbt!]
\caption{ 
MultiSoc model experiments for human policy robustness. Grey cells are variables variability, better results are highlighted (in bold), $R$ and $H$ stand resp. for number of Robots and Humans, $H-Policy$ is the human navigation policy during tests (ORCA is still used during training). Tests are done on $N_{test}$ episodes of 150 timesteps each with random seeds. Thus $150k$ test timesteps means $N_{test}=1000$. 
}
\label{tab:results}
\begin{center}
\begin{tabular}{
  |  *{6}{l} |
  *{4}{l}|
  *{4}{l}
}
\toprule
  \multicolumn{6}{|c|}{Metrics} &
  \multicolumn{4}{c|}{Training Conditions} &
  \multicolumn{4}{c}{Test Conditions} \\
  \rotatebox{90}{\shortstack[c]{Success \\ $\rightarrow$}} & \rotatebox{90}{\shortstack[c]{Collision \\ $\leftarrow$}} & \rotatebox{90}{\shortstack[c]{Intrusion \\  $\leftarrow$ Ratio}} & \rotatebox{90}{\shortstack[c]{Travel  \\ $\leftarrow$ Time}} & \rotatebox{90}{\shortstack[c]{Travel \\ $\leftarrow$ Length}} & \rotatebox{90}{\shortstack[c]{Reward\\ $\rightarrow$}} & \rotatebox{90}{\shortstack[c]{N$_{head}$}} & \rotatebox{90}{\shortstack[c]{N$_{train}$\\ timesteps}} & R & H & \rotatebox{90}{\shortstack[c]{Test\\ timesteps}} & R & H & H-Policy \\

\hline
   0.96 & 0.03 & 4.33 & 14.70 & 14.79 & 19.07 & \multirow{3}{*}{4} & \multirow{3}{*}{20M} & \multirow{3}{*}{1} & \multirow{3}{*}{20} & \multirow{3}{*}{150k} & \multirow{3}{*}{1} & \multirow{3}{*}{20} & \cellcolor{ashgrey!25} ORCA\\
  \textbf{0.98} & \textbf{0.01} & \textbf{3.66} & 15.31 & 15.41 & \textbf{19.47} & & & & & & & &\cellcolor{ashgrey!25} ORCA+SF\\
  0.69 & 0.31 & 9.17 & \textbf{13.12} & \textbf{11.28} & 13.71 & & & & & & & & \cellcolor{ashgrey!25} ORCA+FoV\\
 \hline

   0.89 & 0.08 & \textbf{6.79} & 14.22 & 13.68 & 69.98  & \multirow{3}{*}{4} & \multirow{3}{*}{20M} & \multirow{3}{*}{5} & \multirow{3}{*}{20} & \multirow{3}{*}{150k} & \multirow{3}{*}{5} & \multirow{3}{*}{20} & \cellcolor{ashgrey!25} ORCA\\
  \textbf{0.92} & \textbf{0.06} & 7.05 & 14.58 & 14.11 & \textbf{74.17} & & & & & & & &\cellcolor{ashgrey!25} ORCA+SF\\
  0.67 & 0.31 & 6.8 & \textbf{13.41 }& \textbf{11.29} & 27.27 & & & & & & & & \cellcolor{ashgrey!25} ORCA+FoV\\
 \hline
\end{tabular}

\end{center}

\end{table*}

\end{document}